\documentclass[letterpaper, 10 pt, conference]{ieeeconf}
\IEEEoverridecommandlockouts
\usepackage{graphicx}
\usepackage{amsmath,amssymb}
\usepackage{booktabs,array,tabularx}
\usepackage[T1]{fontenc}
\usepackage{microtype}
\usepackage{caption}
\usepackage{balance}
\newcolumntype{Y}{>{\raggedright\arraybackslash}X}
\let\reportitemize\itemize
\let\endreportitemize\enditemize
\renewenvironment{itemize}{\reportitemize
  \setlength{\itemsep}{2pt}\setlength{\parsep}{0pt}}{\endreportitemize}
\usepackage{threeparttable}
\usepackage{placeins}
\usepackage{xcolor}
\usepackage{colortbl}
\definecolor{llmrow}{HTML}{EAF1F8}
\usepackage{url}
\usepackage{cite}
\makeatletter
\let\NAT@parse\undefined
\makeatother
\usepackage{hyperref}
\hypersetup{colorlinks=true,
linkcolor=blue,
anchorcolor=blue,
citecolor=blue,
urlcolor=cyan,
breaklinks=true
}

\newcommand{\reporttabsetup}{%
  \renewcommand{\arraystretch}{1.18}%
  \setlength{\tabcolsep}{4pt}%
}
\makeatletter
\renewcommand{\paragraph}{\@startsection{paragraph}{4}{\z@}
  {0.7ex plus 0.3ex minus 0.2ex}{-0.7em}{\normalfont\normalsize\bfseries}}
\makeatother
\newcommand{\placeholderfigure}[2]{%
  \begin{figure}[htbp]
  \centering
  \fbox{\parbox[c][#1][c]{\dimexpr\columnwidth-2\fboxsep-2\fboxrule\relax}{%
    \centering\small\color{black!60}#2}}
}

\title{\LARGE \bf
Frontier LLMs Enter the World of Robotics:\\
A Cross-Model Evaluation on RoboColiseum
}

\author{Da Huang$^{*}$, Jichao Wang$^{*}$, Junhui Wu$^{*}$, Tong Ran$^{*}$, Linting Guo, Siying Zhang,\\
Chuang Men, Sukai Wang, Yue Zhou, Maoqing Yao$^{\dagger}$, and Mo Wu$^{\dagger}$%
\thanks{$^{*}$Equal contribution. $^{\dagger}$Corresponding authors.}%
}

\begin{document}
\maketitle
\thispagestyle{plain}
\pagestyle{plain}

\begin{abstract}
Recent studies have explored the potential of multimodal large language models for robot control, yet the models' embodied capabilities remain insufficiently characterized across models, tasks, and execution settings. We conduct a multidimensional comparative evaluation on RoboColiseum across four tracks: Instruction Following, Spatial Reasoning, General Manipulation, and Robustness. We implement Agent2Act, a control harness that integrates GPT-6 Astra, GPT-6.1 Sol, Claude Opus 5.5, and Gemini 4 Argon into the evaluation environment through shared observation, action execution, and feedback interfaces. Under common task settings and scoring criteria, all four models exhibit a consistent capability profile: high instruction-following scores, with three approaching perfect performance; strong spatial reasoning; and general manipulation scores below those of the other three tracks. Within the evaluated configurations, Astra leads instruction following and robustness, while Opus leads spatial reasoning, general manipulation, and overall performance. These results reveal complementary strengths across models and a persistent gap between task understanding and reliable physical execution. Component comparisons on selected tasks further show that harness design affects task success, planning cost, and execution time. Additionally, case studies with Astra examine whole-body coordination, manipulation across robot embodiments, precision manipulation, dexterous-hand manipulation, and real-robot tasks, providing further insight into the model's capabilities and limitations across embodied tasks.
\end{abstract}

\section{Introduction}
\label{sec:introduction}
\label{sec:overview}
\subsection{Motivation}
The release of GPT-6 Astra~\cite{openai2026astra} raises a practical question: what can a
multimodal large language model achieve when it directly controls a robot?
A growing body of work explores how large language models  (LLMs) can use
observations to control robots.

In simulation, W. Zhang et al. evaluate LLMs as manipulation policies and
identify a gap between semantic task understanding and precise physical
control \cite{zhang2026unexpected}. GPT-as-Policy compares Astra-generated
end-effector actions with a hybrid controller in which Astra reviews and
corrects $\pi_{0.5}$ action proposals \cite{su2026astra}. Galbot's study
examines direct and hybrid control across simulated manipulation and other
embodied domains, distinguishing model-generated commands from actions
provided by learned policies or whole-body controllers
\cite{galbot2026embodied}. Y. Yao et al. evaluate tabletop and mobile
manipulation in Ebench, where Astra generates end-effector waypoints,
gripper targets, and base targets through a fixed execution interface
\cite{yao2026frontier}.

On real hardware, RoboCurve compares Astra with a VLA policy on bimanual
tabletop tasks using YAM robot arms. Astra issues absolute end-effector
poses through action tools \cite{zhang2026robocurve}. RoboICL evaluates
both simulated tasks and three real-robot tasks on a Franka Research 3
arm. Its frozen Astra model uses demonstrations and interaction memory
to generate numerical action chunks through a fixed tool schema
\cite{liu2026roboicl}.

These studies provide initial evidence of LLM-driven robot control across
different tasks and control configurations. Building on this work, we
compare four frontier multimodal large language models across
RoboColiseum's four tracks, using shared sensing and execution interfaces
to examine their strengths and limitations.

Our evaluation covers GPT-6 Astra~\cite{openai2026astra}, GPT-6.1
Sol~\cite{openai2026sol}, Claude Opus 5.5~\cite{anthropic2026opus}, and
Gemini 4 Argon~\cite{google2026argon}. To integrate these models into
RoboColiseum, we implement Agent2Act, a harness that provides a unified
interface for interaction with robotic environments and action execution.
RoboColiseum's consistent task settings and scoring criteria allow us to
systematically examine the models' relative strengths, weaknesses, and
capability boundaries in embodied tasks.

\subsection{Benchmark Overview and Key Findings}
RoboColiseum is a multidimensional benchmark for embodied model
capabilities. It supports vision-language-action (VLA) models, world/action
models (WAM), and agent-based systems. Its four tracks assess complementary
dimensions of embodied performance:
\begin{itemize}
\item \textbf{Instruction Following:} evaluates instruction understanding
across ten dimensions, including color, size, quantity, semantics, and
logical conditions.
\item \textbf{Spatial Reasoning:} evaluates understanding of spatial
relations, including relative and absolute positions, sorting, and stacking.
\item \textbf{General Manipulation:} evaluates a range of manipulation
skills, including grasping, placing, insertion, twisting, pouring,
scooping, and bimanual coordination.
\item \textbf{Robustness:} evaluates performance under changes in robot
state, camera mounting position, image quality, instruction wording,
environmental conditions, and lighting.
\end{itemize}
Together, these tracks reveal model strengths, limitations, and challenges
in physical task execution. Our main findings are as follows:
\begin{itemize}
\item \textbf{Strong instruction following, spatial reasoning, and
robustness, but weaker manipulation.} All four models exhibit a common
capability profile: their instruction following, spatial reasoning, and
robustness scores exceed their general manipulation scores. Strong
zero-shot instruction following has not yet translated into reliable
complex physical manipulation.

\item \textbf{Different models have different strengths.} Among the four
multimodal models, Astra leads instruction following and robustness,
while Opus leads spatial reasoning, general manipulation, and Overall.
No single model leads every dimension. These complementary strengths
show why Overall scores alone cannot fully characterize model capabilities.

\item \textbf{Cost and latency remain bottlenecks.} Despite strong
benchmark performance, high inference costs and long waits for model
responses continue to limit the practical use of multimodal models for
embodied robot control.

\item \textbf{Harness design matters.} A well-designed harness helps
models make full use of their capabilities in robot control. Agent2Act
provides a shared control harness for evaluation across four capability
dimensions. Comparisons on a limited test set illustrate how observation
inputs, feedback, history, and control interfaces affect task completion
and execution efficiency.

\item \textbf{Case studies reveal the scope and limits of model
capabilities.} Case studies with GPT-6 Astra demonstrate that multimodal
large language models can control whole-body motion, manipulate objects
across robot embodiments, combine mobile navigation with interaction,
and perform selected tasks on real robots. Dexterous-hand control and
fine-grained physical interaction remain challenging.
\end{itemize}

\begin{table}[t]
\centering
\begin{threeparttable}
\caption{Four-model comparison: Overall and four track scores (\%), ordered by Overall.}
\label{tab:intro-three-models}
\label{tab:main-boards}
\reporttabsetup
\small
\begin{tabular*}{\columnwidth}{@{\extracolsep{\fill}}lrrrrr@{}}
\toprule
\textbf{Model} & \textbf{overall} & \textbf{instr} & \textbf{spatial} & \textbf{manip} & \textbf{robust} \\
\midrule
Claude Opus 5.5 & \textbf{76.28} & 96.60 & \textbf{82.50} & \textbf{46.70} & 79.30 \\
GPT-6 Astra & 75.99 & \textbf{99.35} & 78.33 & 45.86 & \textbf{80.42} \\
GPT-6.1 Sol & 70.91 & 95.50 & 77.71 & 40.01 & 70.40 \\
Gemini 4 Argon & 66.86 & 85.89 & 69.38 & 38.79 & 73.36 \\
\bottomrule
\end{tabular*}
\begin{tablenotes}[flushleft]
\footnotesize
\item Bold values denote the highest reported score in each column. Astra, Sol, and Opus use \texttt{high} reasoning effort; Argon uses \texttt{medium}.
\end{tablenotes}
\end{threeparttable}
\end{table}
\begin{figure*}[!t]
    \centering
    \includegraphics[width=\textwidth]{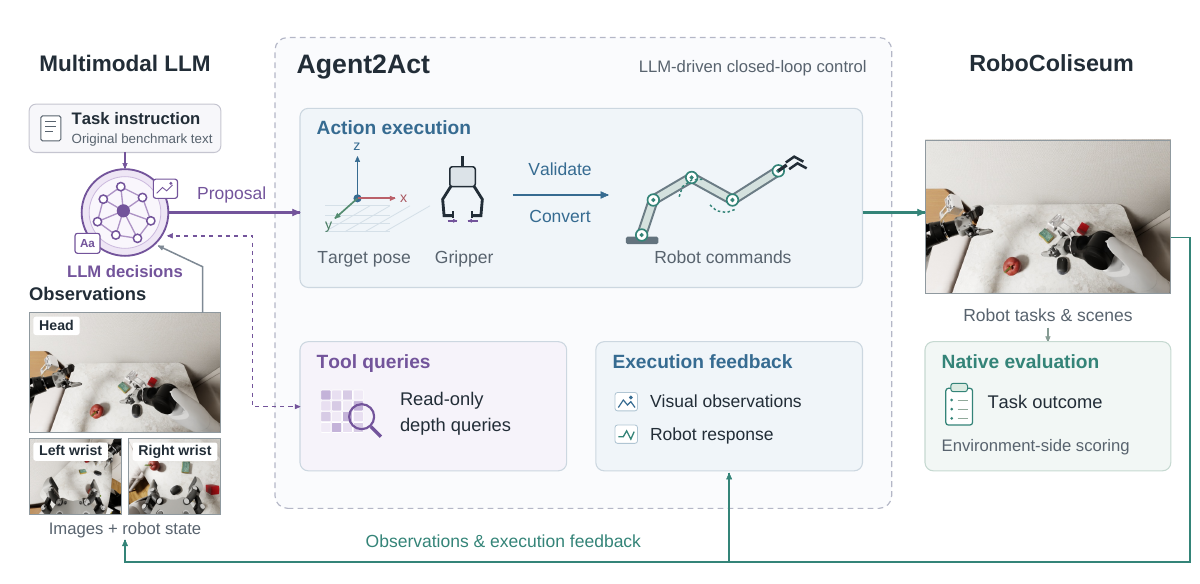}
    \caption{Agent2Act's LLM-driven control loop. The multimodal
    LLM uses task context and observations, queries read-only
    sensor evidence, and proposes actions. Agent2Act validates
    and converts proposals into robot commands; subsequent
    observations, robot responses, and task progress inform the
    next decision. RoboColiseum determines task outcomes through
    native evaluation. Camera panels show recorded observations.}
    \label{fig:agent2act-framework}
\end{figure*}

\section{Agent2Act Harness}
\label{sec:method}

\subsection{Framework Workflow}

Agent2Act connects a multimodal large language model (LLM) to a
robot through a closed loop of observation, decision, execution,
and feedback (Figure~\ref{fig:agent2act-framework}). In our
experiments, the model interprets the task and scene, selects the
next manipulation objective, and specifies motion targets and
gripper actions. Agent2Act provides the model with sensing,
execution, and feedback interfaces and translates its decisions
into physical interactions in the simulator.

At each turn, the model receives current visual observations,
robot state, and a concise history of previous decisions and their
observed effects. It uses this information to advance the task,
assess earlier actions, and revise its plan. Direct control therefore
means that the LLM determines which action to attempt, while the
execution layer converts that decision into robot commands.

\subsection{Model Input Modules}
\label{sec:model-inputs}
Table~\ref{tab:model-input-modules} summarizes the contents and
roles of the model input modules. The same Agent2Act harness
constructs these inputs for all four models.

\begin{table*}[t]
\centering
\begin{threeparttable}
\caption{Model input modules and their use in the reported setting.}
\label{tab:model-input-modules}
\reporttabsetup\small
\begin{tabularx}{\textwidth}{@{}>{\raggedright\arraybackslash}p{0.18\textwidth}>{\raggedright\arraybackslash}p{0.46\textwidth}Y@{}}
\toprule
\textbf{Module} & \textbf{Information actually supplied to the model} & \textbf{Use and scope} \\
\midrule
Original task instruction & The benchmark's original natural-language request, including the requested object, arm, relation, or operation. & Provided as the task instruction; kept separate from authored guidance. \\
General robot instructions & Action schema, coordinate and pose conventions, robot/gripper geometry, available controls, camera calibration, and execution constraints. & Shared control interface; No geometry and No projection remove the specified information or requirement. \\
Authored operation guidance & Optional task-specific requirements and general manipulation suggestions, such as contact or operation strategies. & Configurable input, separate from the original instruction; not supplied in the benchmark evaluation on any of the four tracks. \\
Visual and measured state & Current RGB observations, measured robot poses/joints and gripper state, plus depth observations with simulated noise through the configured input/query interface. & Supplied observations; No depth removes depth inputs and queries. \\
Execution feedback & Requested-versus-observed motion summaries and action-validation/rejection information. & Supports correction; No feedback removes deviation summaries while retaining current observations. \\
Recent history & Previous decisions, queries, execution records, and configured observation history. & Supplied in the full setting; removed in the No history comparison. \\
\bottomrule
\end{tabularx}
\end{threeparttable}
\end{table*}

Authored guidance is an optional framework capability separate from
the original benchmark instruction. It is not supplied in the reported
benchmark evaluation: all tasks in all four tracks use the original
benchmark instructions without supplementary task prompts.
General robot control instructions remain part of the shared interface.

The LLM decides when to query depth based on the current
observation. In simulation, the tool returns depth observations
with simulated noise rather than ground-truth values. These
read-only queries do not execute robot actions.

\subsection{Action Validation, Execution, and Feedback}

The LLM expresses each decision as a manipulation objective,
candidate motion targets, gripper commands, and an expected
observable outcome. Motion targets use a common robot reference
frame, and Agent2Act converts them into executable robot commands.

Agent2Act checks candidates for kinematic feasibility and
compliance with the action interface, then executes the
highest-ranked admissible candidate. If no candidate passes
these checks, it returns the rejection reasons to the model
for revision. Actions execute over a bounded horizon before
a new observation returns control to the model, allowing
subsequent decisions to respond to the actual robot and scene state.

The next observation closes the loop with visual evidence and
the measured robot response. The LLM compares this feedback
with the intended outcome to decide whether to continue,
correct its action, or declare completion. Agent2Act records
model-proposed actions, issued commands, and observed execution
outcomes separately: generating an executable command does not
establish that the robot has completed the operation.

\subsection{Benchmark Integration}

RoboColiseum supplies the scenes, robots, task instructions,
action budgets, and scoring rules. Agent2Act connects the model
to this environment while preserving the original task definitions
and evaluation protocol. Section~\ref{sec:model-inputs} details the
model inputs used in the reported experiments. The benchmark's
native evaluation determines final task outcomes, independently
of the model's completion judgment.   
\section{Benchmark Evaluation Results}
\label{sec:evaluation}

\subsection{Evaluation Protocol}
We report the final results for GPT-6 Astra, GPT-6.1 Sol, Claude
Opus 5.5, and Gemini 4 Argon on RoboColiseum's Instruction Following,
Spatial Reasoning, General Manipulation, and Robustness tracks.

The evaluation follows the official RoboColiseum benchmark configuration
and scoring protocol. Each task score is the mean of its substep scores,
and each track score is the mean of its constituent task scores. Tasks
are weighted equally rather than by their numbers of cases. The main
evaluation uses the benchmark's native substep scores.

Each model is evaluated once per case. For corresponding cases in each
track, all four models use identical samples, the same fixed random seed,
and a shared Agent2Act harness. GPT-6 Astra, GPT-6.1 Sol, and Claude
Opus 5.5 use \texttt{high} reasoning effort. Gemini 4 Argon uses
\texttt{medium} because responses through the API access channel used
in this evaluation were too slow at \texttt{high} for large-scale batch
evaluation. Comparisons involving Argon therefore reflect differences
in both the model and its reasoning setting.

\begin{figure}[!htbp]
\centering
\includegraphics[width=\textwidth]{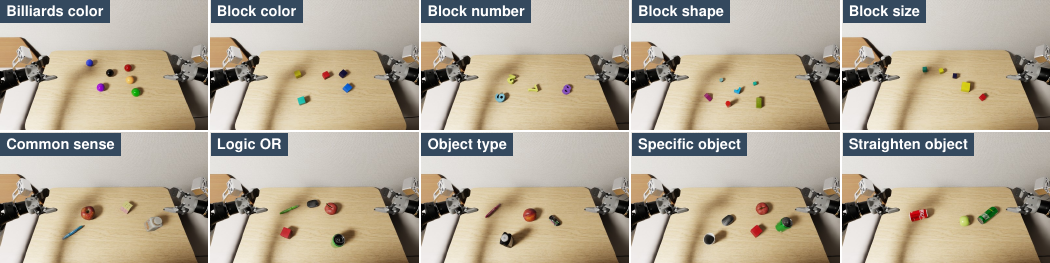}
\par\vspace{2pt}
{\fontsize{7}{9}\selectfont\textrm{(a) Instruction Following}\par}
\vspace{6pt}
\includegraphics[width=\textwidth]{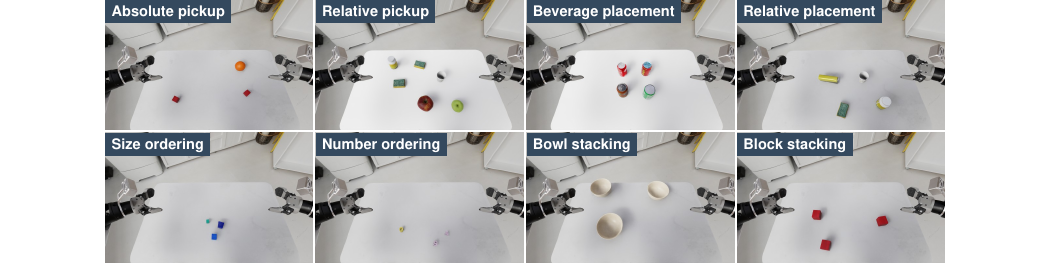}
\par\vspace{2pt}
{\fontsize{7}{9}\selectfont\textrm{(b) Spatial Reasoning}\par}
\vspace{6pt}
\includegraphics[width=\textwidth]{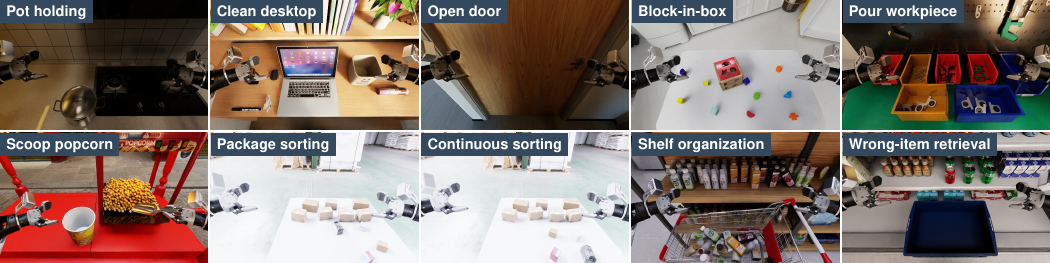}
\par\vspace{2pt}
{\fontsize{7}{9}\selectfont\textrm{(c) General Manipulation}\par}
\vspace{6pt}
\includegraphics[width=\textwidth]{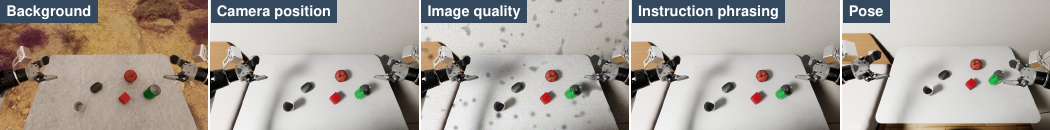}
\par\vspace{2pt}
{\fontsize{7}{9}\selectfont\textrm{(d) Robustness}\par}
\caption{Rendered scene examples from the four RoboColiseum tracks:
(a) Instruction Following, (b) Spatial Reasoning, (c) General Manipulation,
and (d) Robustness. The Robustness examples show the same object-selection
task under five perturbation types. These native camera previews illustrate
the task settings and do not depict model execution outcomes.}
\label{fig:benchmark-track-examples}
\end{figure}

All tasks in the four tracks use the original benchmark instructions,
without supplementary task-specific prompts. General robot instructions
and control-interface descriptions are identical across the four models.
Appendix~\ref{app:benchmark-renderings} provides examples of rendered
task scenes from all four tracks, including the five Robustness
perturbation types (Fig.~\ref{fig:benchmark-track-examples}).

\subsection{Model Leaderboard}
Table~\ref{tab:benchmark-model-comparison} compares the four multimodal
large language models evaluated here with leading policy models from
the official Overall leaderboard. Models are ordered by reported Overall
score, with columns showing Overall followed by the four track scores.

\begin{table*}[t]
\centering
\begin{threeparttable}
\caption{Model leaderboard: Overall and four track scores on a 0--1 scale.}
\label{tab:benchmark-model-comparison}
\reporttabsetup\small
\begin{tabularx}{\textwidth}{@{}r>{\raggedright\arraybackslash}p{0.30\textwidth}*{5}{>{\raggedleft\arraybackslash}X}@{}}
\toprule
\textbf{Rank} & \textbf{Model} & \textbf{Overall} & \shortstack{\textbf{Instruction}\\\textbf{Following}} & \shortstack{\textbf{Spatial}\\\textbf{Reasoning}} & \shortstack{\textbf{General}\\\textbf{Manipulation}} & \textbf{Robustness} \\
\midrule
1 & Paimon & \textbf{0.7754} & 0.8961 & 0.6791 & \textbf{0.6613} & \textbf{0.8653} \\
2 & HiDream-O1-Embodied & 0.7692 & 0.8913 & 0.7437 & 0.6442 & 0.7978 \\
\rowcolor{llmrow}
3 & \textbf{Claude Opus 5.5} & 0.7628 & 0.9660 & \textbf{0.8250} & 0.4670 & 0.7930 \\
\rowcolor{llmrow}
4 & \textbf{GPT-6 Astra} & 0.7599 & \textbf{0.9935} & 0.7833 & 0.4586 & 0.8042 \\
\rowcolor{llmrow}
5 & \textbf{GPT-6.1 Sol} & 0.7091 & 0.9550 & 0.7771 & 0.4001 & 0.7040 \\
6 & DM0.5 & 0.7076 & 0.8444 & 0.6146 & 0.6370 & 0.7344 \\
\rowcolor{llmrow}
7 & \textbf{Gemini 4 Argon} & 0.6686 & 0.8589 & 0.6938 & 0.3879 & 0.7336 \\
8 & GigaBrain-0.7 & 0.6504 & 0.8166 & 0.4729 & 0.6320 & 0.6800 \\
\bottomrule
\end{tabularx}
\begin{tablenotes}[flushleft]
\footnotesize
\item Light blue rows denote multimodal models evaluated through Agent2Act; bold numeric values denote the highest reported score in each column. External entries come from the \href{https://robocoliseum.ai/leaderboard?metric=overall}{official RoboColiseum Overall leaderboard} and retain their respective training and execution settings.
\end{tablenotes}
\end{threeparttable}
\end{table*}

Among the four multimodal models, Opus leads Overall, Spatial Reasoning,
and General Manipulation, while Astra leads Instruction Following and
Robustness. Opus and Astra rank third and fourth in this leaderboard,
separated by 0.0029 in Overall; Sol ranks fifth. All three exceed the
highest policy score listed in the table on both Instruction Following
and Spatial Reasoning, demonstrating strong instruction understanding
and spatial reasoning alongside competitive overall performance.

Argon, evaluated at \texttt{medium} reasoning effort, ranks seventh
with an Overall score of 0.6686. Its Instruction Following and Spatial
Reasoning scores are lower than those of the other three multimodal
models, but its Robustness score exceeds Sol's (0.7336 versus 0.7040).
These differences highlight the multidimensional nature of model
capabilities and the importance of examining individual dimensions
rather than Overall alone.

All four models score higher on Instruction Following and Spatial
Reasoning than on General Manipulation. They also trail every listed
policy on General Manipulation, with scores of
0.3879--0.4670, compared with 0.6320--0.6613 for the policies.
\textbf{These results identify general manipulation as a shared weakness
of current multimodal large language models: strong instruction
understanding and spatial reasoning have not yet translated into
reliable complex physical manipulation.}

Opus achieves the highest Overall score among the four multimodal models
while consuming the fewest tokens in total. Astra and Sol have similar
total usage, with Sol consuming slightly more, while Argon consumes the
most. At the reference API prices, Sol achieves a score close to Astra's
at a substantially lower estimated inference cost, offering a favorable
balance of performance and cost. Opus scores slightly higher than Astra
while using fewer tokens at lower unit prices. Argon's estimated
inference cost is slightly higher than Sol's.

\subsection{Task-Level Performance}
Tables~\ref{tab:main-instruction}--\ref{tab:main-manip} report task scores
for Instruction Following, Spatial Reasoning, and General Manipulation
on a 0--1 scale, rounded to two decimal places.
Table~\ref{tab:robust-dimensions} summarizes Robustness by perturbation
type. Bold values identify the highest reported score in each row.
All four tracks follow the shared instruction setting described above.

Astra, Sol, and Opus achieve near-perfect Instruction Following scores
overall, while Argon's performance varies more across tasks. Argon scores
only 0.60 on object straightening, compared with 0.98 for Astra, 0.84 for
Sol, and 0.83 for Opus. This task requires both picking up the target
object and rotating the wrist to straighten it. The additional
manipulation requirement may explain the lower scores and failed attempts.

\begin{table*}[t]
\centering
\begin{threeparttable}
\caption{Instruction Following: reported score for each task on a 0--1 scale.}
\label{tab:main-instruction}
\reporttabsetup\small
\begin{tabular*}{\textwidth}{@{\extracolsep{\fill}}lrrrr@{}}
\toprule
\textbf{Task} & \textbf{GPT-6 Astra} & \textbf{GPT-6.1 Sol} & \textbf{Claude Opus 5.5} & \textbf{Gemini 4 Argon} \\
\midrule
Billiards color & \textbf{1.00} & 0.98 & 0.99 & 0.87 \\
Block color & \textbf{1.00} & 0.99 & \textbf{1.00} & 0.90 \\
Block number & \textbf{1.00} & \textbf{1.00} & \textbf{1.00} & 0.94 \\
Block shape & \textbf{0.99} & 0.91 & 0.91 & 0.85 \\
Block size & \textbf{0.98} & 0.95 & 0.95 & 0.90 \\
Common sense & 0.99 & 0.96 & \textbf{1.00} & 0.89 \\
Logic OR & \textbf{1.00} & 0.99 & \textbf{1.00} & 0.93 \\
Object type & \textbf{1.00} & 0.96 & 0.99 & 0.85 \\
Specific object & \textbf{1.00} & 0.98 & 1.00 & 0.87 \\
Straighten object & \textbf{0.98} & 0.84 & 0.83 & 0.60 \\
\bottomrule
\end{tabular*}
\end{threeparttable}
\end{table*}

All four models show strong Spatial Reasoning performance overall,
with differences across individual tasks. Opus scores 1.00 on both
position-based pickup tasks, while Sol scores 1.00 on size ordering
and block stacking. Argon scores 0.88 on absolute pickup and 0.77 on
number ordering, exceeding Sol's respective scores of 0.83 and 0.60,
but trails the other models on most spatial tasks. All four models
score zero on bowl stacking, primarily because repeated unsuccessful
grasps at the bowl rim exhaust the step budget.

\begin{table*}[t]
\centering
\begin{threeparttable}
\caption{Spatial Reasoning: reported score for each task on a 0--1 scale.}
\label{tab:main-spatial}
\reporttabsetup\small
\begin{tabular*}{\textwidth}{@{\extracolsep{\fill}}lrrrr@{}}
\toprule
\textbf{Task} & \textbf{GPT-6 Astra} & \textbf{GPT-6.1 Sol} & \textbf{Claude Opus 5.5} & \textbf{Gemini 4 Argon} \\
\midrule
Absolute pickup & 0.92 & 0.83 & \textbf{1.00} & 0.88 \\
Relative pickup & 0.93 & 0.97 & \textbf{1.00} & 0.80 \\
Beverage placement & 0.93 & 0.88 & \textbf{0.98} & 0.85 \\
Relative placement & 0.85 & \textbf{0.93} & 0.92 & 0.82 \\
Size ordering & 0.90 & \textbf{1.00} & 0.93 & 0.77 \\
Number ordering & 0.87 & 0.60 & \textbf{0.90} & 0.77 \\
Bowl stacking & \textbf{0.00} & \textbf{0.00} & \textbf{0.00} & \textbf{0.00} \\
Block stacking & 0.87 & \textbf{1.00} & 0.87 & 0.67 \\
\bottomrule
\end{tabular*}
\end{threeparttable}
\end{table*}

General Manipulation remains the weakest track for all four models.
Our analysis of the recorded runs indicates that most failures stem
from limited understanding of object affordances. The pot-holding case
in Section~\ref{sec:case-instruction-detail} illustrates how explicit
interaction guidance improves grasping, highlighting the difficulty
of inferring suitable object-interaction strategies from task
instructions alone. Opus leads pot holding at 0.60, while Argon scores
0.00. Argon leads package sorting at 0.56, and Astra leads door opening
and shelf organization at 0.86 and 0.67, respectively.

\begin{table*}[t]
\centering
\begin{threeparttable}
\caption{General Manipulation: reported score for each task on a 0--1 scale.}
\label{tab:main-manip}
\reporttabsetup\small
\begin{tabular*}{\textwidth}{@{\extracolsep{\fill}}lrrrr@{}}
\toprule
\textbf{Task} & \textbf{GPT-6 Astra} & \textbf{GPT-6.1 Sol} & \textbf{Claude Opus 5.5} & \textbf{Gemini 4 Argon} \\
\midrule
Clean desktop & 0.07 & 0.05 & \textbf{0.20} & 0.18 \\
Pot holding & 0.41 & 0.23 & \textbf{0.60} & 0.00 \\
Open door & \textbf{0.86} & 0.81 & 0.68 & 0.81 \\
Block-in-box & \textbf{0.83} & 0.80 & 0.81 & 0.75 \\
Pour workpiece & 0.11 & 0.10 & \textbf{0.24} & 0.16 \\
Scoop popcorn & \textbf{0.43} & 0.36 & 0.29 & 0.20 \\
Package sorting & 0.39 & 0.41 & 0.54 & \textbf{0.56} \\
Continuous sorting & 0.00 & 0.00 & 0.05 & \textbf{0.07} \\
Shelf organization & \textbf{0.67} & 0.49 & 0.49 & 0.41 \\
Wrong-item retrieval & \textbf{0.83} & 0.77 & 0.79 & 0.74 \\
\bottomrule
\end{tabular*}
\end{threeparttable}
\end{table*}

\subsection{Performance Across Perturbation Types}
\label{sec:robust-dimensions}
We report scores under five perturbation conditions: background and
lighting, camera position, image quality, instruction phrasing, and
initial pose. The perturbation test set is derived from the Instruction
Following test set by applying perturbations independently along each
dimension. The perturbed samples and IF baseline share the same
underlying tasks, layouts, original instructions, and aggregation
weights; only the specified perturbation dimension is varied.
Instruction-phrasing perturbations change the wording while preserving
the original task intent. Table~\ref{tab:robust-dimensions} compares
each condition with the model's own IF baseline.

\begin{table*}[t]
\centering
\begin{threeparttable}
\caption{Robustness by perturbation type, compared with each model's Instruction Following (IF) baseline. Scores and $\Delta$ are on a 0--1 scale, rounded to two decimal places; $\Delta$ is the IF baseline minus the perturbation score.}
\label{tab:robust-dimensions}
\reporttabsetup\small
\begin{tabular*}{\textwidth}{@{\extracolsep{\fill}}lrrrrrrrr@{}}
\toprule
& \multicolumn{2}{c}{\textbf{GPT-6 Astra}} & \multicolumn{2}{c}{\textbf{GPT-6.1 Sol}} & \multicolumn{2}{c}{\textbf{Claude Opus 5.5}} & \multicolumn{2}{c}{\textbf{Gemini 4 Argon}} \\
\cmidrule(lr){2-3}\cmidrule(lr){4-5}\cmidrule(lr){6-7}\cmidrule(l){8-9}
\textbf{Setting} & \textbf{Score} & $\Delta$ & \textbf{Score} & $\Delta$ & \textbf{Score} & $\Delta$ & \textbf{Score} & $\Delta$ \\
\midrule
IF baseline & 0.99 & --- & 0.96 & --- & 0.97 & --- & 0.86 & --- \\
\midrule
Background and lighting & 0.84 & 0.16 & 0.80 & 0.16 & 0.81 & 0.16 & 0.78 & 0.08 \\
Camera position & 0.85 & 0.14 & 0.86 & 0.10 & 0.77 & 0.20 & 0.72 & 0.14 \\
Image quality & 0.68 & \textbf{0.32} & 0.18 & \textbf{0.77} & 0.76 & \textbf{0.20} & 0.60 & \textbf{0.26} \\
Instruction phrasing & 0.86 & 0.13 & 0.85 & 0.11 & 0.79 & 0.18 & 0.77 & 0.09 \\
Initial pose & 0.79 & 0.20 & 0.83 & 0.12 & 0.84 & 0.13 & 0.80 & 0.06 \\
\bottomrule
\end{tabular*}
\begin{tablenotes}[flushleft]\footnotesize
\item Bold identifies the largest baseline gap for each model across the five perturbation types. Gaps are calculated before rounding.
\end{tablenotes}
\end{threeparttable}
\end{table*}

Image-quality perturbations include random frame loss across the head
and two wrist cameras, lens occlusion, and different types of image
noise. Among the tested perturbation types and intensities, image-quality
degradation produces the largest drop from each model's own baseline:
0.77 for Sol, 0.32 for Astra, 0.26 for Argon, and 0.20 for Opus.
Sol is the most sensitive to this condition. Opus's score drops vary least across the five dimensions, and it leads under
image-quality perturbations with a score of 0.76.

Among the tested conditions, the smallest drops occur under instruction
phrasing for Astra (0.13), camera position for Sol (0.10), and initial
pose for Opus (0.13) and Argon (0.06).

Despite its lower Instruction Following baseline, Argon remains
competitive with Sol under perturbations and achieves a higher
Robustness score overall.
\section{Comparison of Harness Components}
\label{sec:experiments}
We evaluate how different functional modules affect the operation of the
harness on a limited test set, and qualitatively analyze their roles using
task outcomes and execution costs.

\subsection{Experimental Setup and Reporting}
\label{sec:ablation-setup}
All experiments use GPT-6 Astra with high reasoning effort and the original benchmark instructions. The shared robot-control instructions and specified configuration changes are stated explicitly. Each baseline--variant comparison uses matched settings and changes only the specified configuration. The full system provides current RGB images, measured robot state, queries of simulated noisy depth, gripper geometry, execution-deviation feedback, and recent history. Table~\ref{tab:ablation-settings} defines the interventions.

\paragraph{Task subsets}
$\mathcal{I}_{10}$ contains ten Instruction Following (IF) tasks, with
fixed layouts and identical instructions across conditions. All comparisons
except history use this set. The history comparison uses $\mathcal{P}_{4}$,
which comprises four layouts of the block-in-box task. Each task or layout
is run once per condition, with a full scene reset before each run.

\begin{table*}[t]
\centering
\begin{threeparttable}
\caption{The seven configuration changes and their purposes.}
\label{tab:ablation-settings}
\reporttabsetup\small
\begin{tabularx}{\textwidth}{@{}>{\raggedright\arraybackslash}p{0.18\textwidth}cY@{}}
\toprule
\textbf{Component} & \textbf{Set} & \textbf{Intervention and purpose} \\
\midrule
Execution feedback & $\mathcal{I}_{10}$ & Remove requested-versus-observed motion summaries; retain current images, measured state, depth, and history. Examine explicit feedback for correction and task progress. \\
Depth queries & $\mathcal{I}_{10}$ & Remove simulated noisy depth inputs and query tools; retain RGB, camera calibration, robot geometry/state, feedback, and history. Examine completion and planning cost without depth. \\
IK checking & $\mathcal{I}_{10}$ & Hard IK rejects targets that fail strict IK validation and requests replanning. Soft IK allows approximate execution when strict pose acceptance is not met. Examine task completion and subsequent motion correction. \\
Camera projection & $\mathcal{I}_{10}$ & Remove the requirement to compute and report the first motion waypoint's image projection for self-checking. Retain RGB, calibration, depth, geometry, and local IK. Examine explicit geometric checking. \\
Gripper geometry & $\mathcal{I}_{10}$ & Remove gripper dimensions, tool axes, asymmetric-finger descriptions, and contact guidance; retain vision, depth, calibration, and the fixed wrist-to-TCP transform. Examine approach orientation, finger-pad adjustment, and grasping. \\
Action representation & $\mathcal{I}_{10}$ & Replace EEF targets and their IK conversion with joint increments planned from measured joints and forward-kinematics poses. Enable both arms and grippers, with waist, head, and base control disabled. \\
History & $\mathcal{P}_{4}$ & Remove past decision, query, and execution records; retain current observations, depth, and execution-deviation feedback. Examine continuity and recovery. \\
\bottomrule
\end{tabularx}
\end{threeparttable}
\end{table*}

\paragraph{Cost measures}
A planning decision is a model action proposal, including a rejected
proposal; issued steps are the action steps sent to the robot. Success and
costs are evaluated up to the same endpoint: episode termination or 30
cumulative local IK rejections, whichever occurs first.
Appendix~\ref{app:ablations} reports task-level IF outcomes
(Table~\ref{tab:ablation-tasks}) and history-comparison outcomes and runtimes
(Table~\ref{tab:history-layouts}).

\subsection{Comparison Results}
\label{sec:ablations}
\begin{figure*}[t]
\centering
\includegraphics[width=\textwidth]{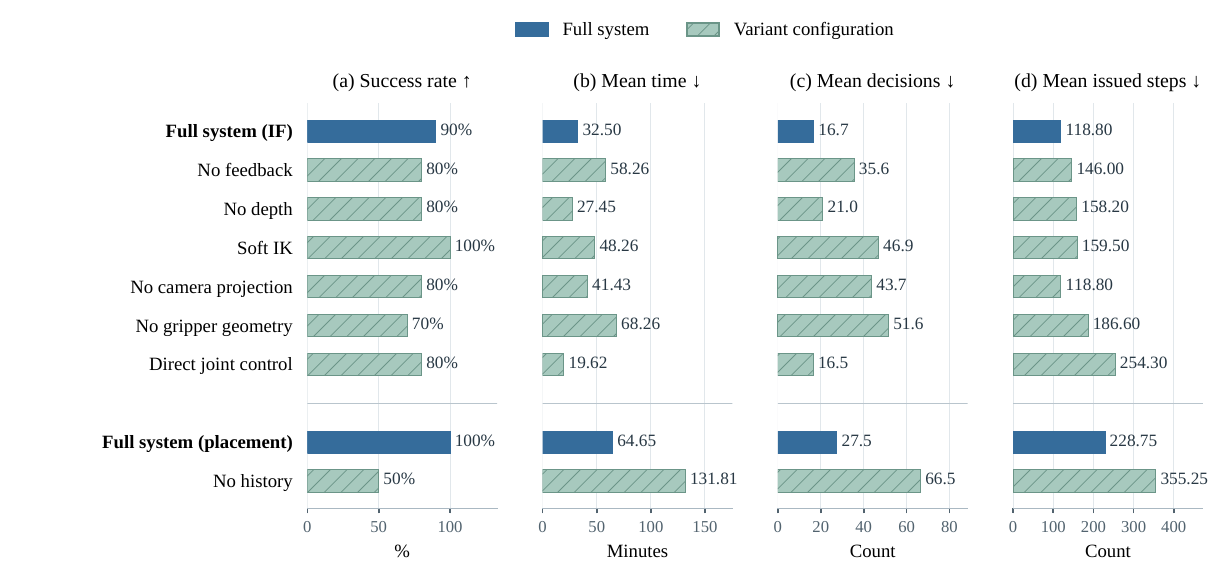}
\caption{Configuration comparison of success rate, mean time, planning decisions,
and issued steps. Solid bars denote the full system; hatched bars denote
variant configurations.}
\label{fig:ablation-comparison}
\end{figure*}

Figure~\ref{fig:ablation-comparison} summarizes success rates and execution
costs. Each comparison follows the termination rule above and considers
task completion alongside execution cost. The comparisons evaluate complete
configurations, including their respective action-validation and execution
mechanisms.

\paragraph{Execution-deviation feedback}
\textbf{Execution-deviation feedback helps the model identify differences between intended and observed motion, reducing unproductive replanning and the associated overhead.}
Without this feedback, the model has more difficulty determining whether
an action achieved its intended effect and may repeat attempts without
effectively correcting the error. On the tested tasks, success decreases
from 90\% to 80\%, while mean planning decisions increase from 16.7 to 35.6.
Mean time rises from 32.50 to 58.26 minutes (+79.3\%), and mean issued steps
increase from 118.8 to 146.0 (+22.9\%). Repeated planning is thus accompanied
by higher runtime and action costs.

\paragraph{Depth queries}
\textbf{Noisy depth observations support target localization and reduce planning decisions and issued steps on the tested tasks.}
The full system queries simulated noisy depth rather than ground-truth
values, as described in Section~\ref{sec:model-inputs}. Without these
inputs and query tools, the model relies on RGB images, camera calibration,
and robot state to estimate three-dimensional target positions. Success
decreases from 90\% to 80\%, mean planning decisions increase from 16.7 to
21.0, and mean issued steps increase from 118.8 to 158.2.

\paragraph{IK checking}
\label{sec:geometry-ablations}
\textbf{Soft IK allows progress through approximate motions, completing more tasks at higher planning, action, and time costs.}
When a target fails strict pose validation, hard IK rejects it for
replanning, whereas Soft IK can execute an approximate solution. The model
then uses subsequent observations to correct the difference between the
requested and executed motion. Soft IK completes all ten tasks, compared
with nine for the full system, but mean planning decisions increase from
16.7 to 46.9 (+180.8\%) and mean time rises from 32.50 to 48.26 minutes.
Mean issued steps also increase from 118.8 to 159.5 (+34.3\%), illustrating
a trade-off between task completion and correction costs.

\paragraph{Explicit camera projection}
\textbf{Removing explicit projection self-checks increases planning decisions and runtime on the tested tasks, while mean issued steps remain unchanged.}
Without the requirement to compute and report the first motion waypoint's
image projection, the model still completes eight of ten tasks using RGB
images, depth queries, and camera calibration, compared with nine for the
full system. The recorded runs include repeated grasp attempts and loss of
grasp. Mean planning decisions increase from 16.7 to 43.7 (+161.7\%), and
mean time rises from 32.50 to 41.43 minutes (+27.5\%).

\paragraph{Gripper geometry}
\textbf{Gripper geometry and contact guidance help the model choose grasping strategies and reduce the cost of repeated attempts.}
Without this information, the model tends to retain its initial gripper
orientation in the recorded grasp attempts, and success decreases from
90\% to 70\%. Mean planning decisions increase from 16.7 to 51.6 (+209.0\%),
mean time rises from 32.50 to 68.26 minutes (+110.0\%), and mean issued
steps increase from 118.8 to 186.6 (+57.1\%). Among the tested IF
configurations, this variant has the lowest success rate and the highest
mean decision count and runtime.

\paragraph{EEF pose control versus direct joint control}
\label{sec:action-representation-ablation}
\textbf{Direct joint control completes tasks but offers a different trade-off among success, issued steps, and runtime compared with EEF pose control.}
Direct joint control completes eight of ten tasks, yielding a slightly
lower success rate than EEF pose control. Mean issued steps increase from
118.8 to 254.3 (+114.1\%). The mean number of planning decisions remains
similar, but the average number of control steps issued per decision
increases substantially. Mean time decreases from 32.50 to 19.62 minutes
(-39.6\%).

\paragraph{History}
\textbf{History helps the model track task progress and use prior attempts to correct errors, reducing repeated actions and the associated overhead.}
Removing history reduces success from 100\% to 50\% across the four
block-in-box layouts. Both failed episodes complete pickup but repeatedly
attempt the same actions during placement. Mean planning decisions increase
from 27.5 to 66.5, mean time rises from 64.65 to 131.81 minutes (+103.9\%),
and mean issued steps increase from 228.75 to 355.25 (+55.3\%). The
layout-level results in Table~\ref{tab:history-layouts} show that the runtime increase is
concentrated in the two failed placement episodes, consistent with the
observed repeated attempts.

\subsection{Summary of Findings}
The comparisons highlight two broad aspects of harness design: information
available to the model affects task completion and the cost of repeated
attempts, while control and validation mechanisms shape trade-offs between
success and execution costs. Planning decisions, issued steps, and runtime
do not always change together, so no single metric fully captures a
configuration's performance. Harness design should therefore be evaluated
jointly in terms of task completion and multiple measures of execution cost.

\section{Case Studies}
\label{sec:cases}
The following cases examine GPT-6 Astra's capabilities in whole-body
control, dexterous-hand control, precision manipulation, operation
across robot embodiments, and long-horizon navigation and manipulation.

We also investigate how supplementary manipulation guidance changes the
model's grasping behavior and improves task success. Finally, we evaluate
the model's manipulation performance on a real robot.

These qualitative cases relate recorded behavior to task outcomes to
characterize demonstrated capabilities and remaining limitations.

\subsection{Whole-Body Control}
Whole-body manipulation combines object handling with changes in robot
position and posture. Here, whole-body control refers to coordination of
the mobile base, torso, and arms. Our interface exposes arm end-effector
targets, torso and head joints, and optional planar base motion. The model
completes trash disposal, shelf-to-table cola transfer, cup retrieval from
a cabinet, door opening and passage, and chair pushing.

\begin{figure}[!tp]
\centering
\includegraphics[width=\columnwidth]{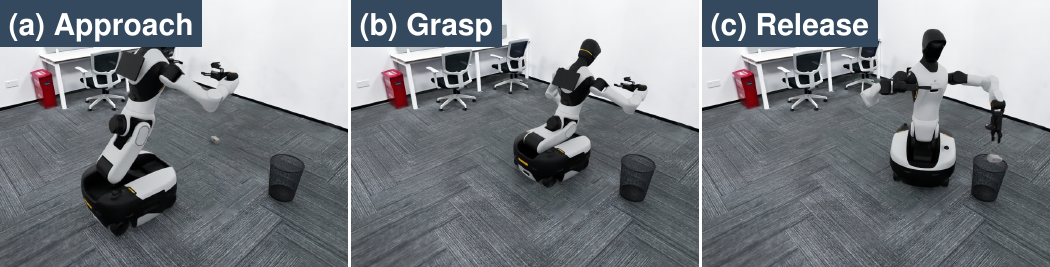}
\caption{Whole-body control during trash disposal, combining base movement, torso adjustment, and arm manipulation.}
\label{fig:case-wbc}
\end{figure}

Trash disposal illustrates how mobility and manipulation work together
(Fig.~\ref{fig:case-wbc}). The model moves the base toward the trash, lowers
the torso to bring the arm within reach, and grasps the object. It then turns
and repositions the base, moves the object above the bin, and releases it.
The episode passes the successive proximity, pickup, destination-proximity,
and inside-bin checks.

These cases show that the model can coordinate whole-body actions through
a common interface to complete multi-stage tasks beyond a fixed tabletop
workspace.

\subsection{Dexterous-Hand Manipulation}
We extend direct joint control to bowl stacking with dexterous hands.
GPT-6 Astra receives visual observations and measured robot state, then
outputs arm targets and 20 finger-joint angle targets per hand to attempt
grasping and stacking.

\begin{figure}[!tp]
\centering
\includegraphics[width=\columnwidth]{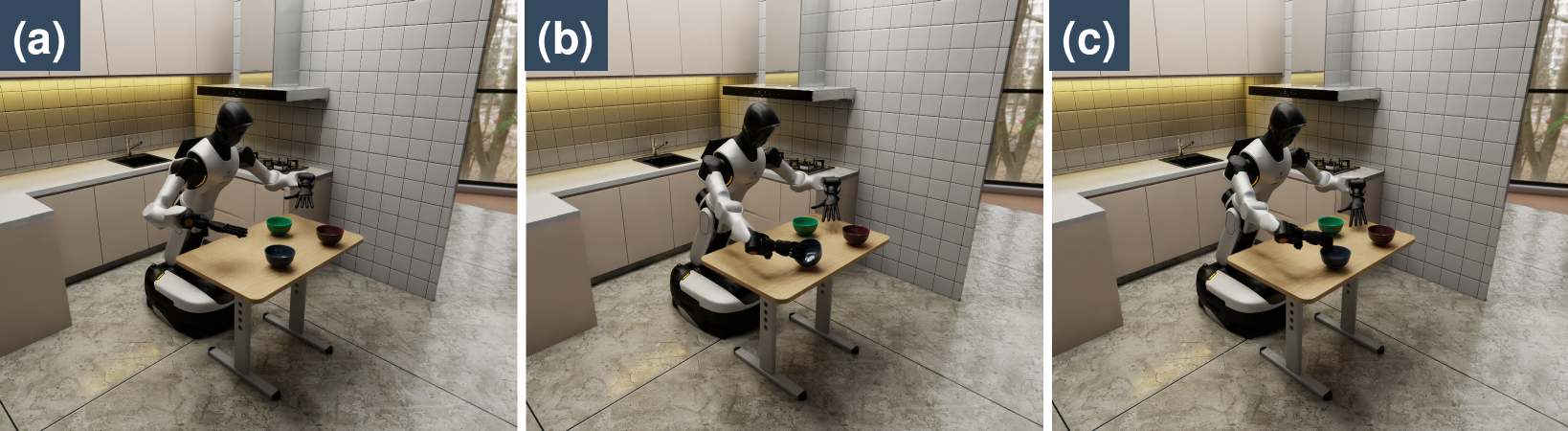}
\caption{Bowl-stacking task where hand and wrist adjustments do not establish a stable grasp.}
\label{fig:case-wuji}
\end{figure}

With measured finger-pad poses, the model approaches the bowl and adjusts
its fingers but does not complete stacking (Fig.~\ref{fig:case-wuji}). A
subsequent configuration additionally provides finger and palm poses,
along with geometric bounding boxes, but still fails to establish a stable
grasp or complete stacking.

In these attempts, finger and wrist adjustments do not produce the
coordinated contacts needed to lift the bowl securely and stack it. Even
with additional hand geometry, the model struggles to control the many
degrees of freedom of a dexterous hand to achieve a stable grasp.

\subsection{Precision Manipulation}
Precision manipulation requires sufficiently accurate end-effector
positions and orientations for alignment. The model can grasp objects
and adjust their poses, but often fails to achieve
the alignment needed for insertion or placement.

\begin{figure}[!tp]
\centering
\includegraphics[width=\columnwidth]{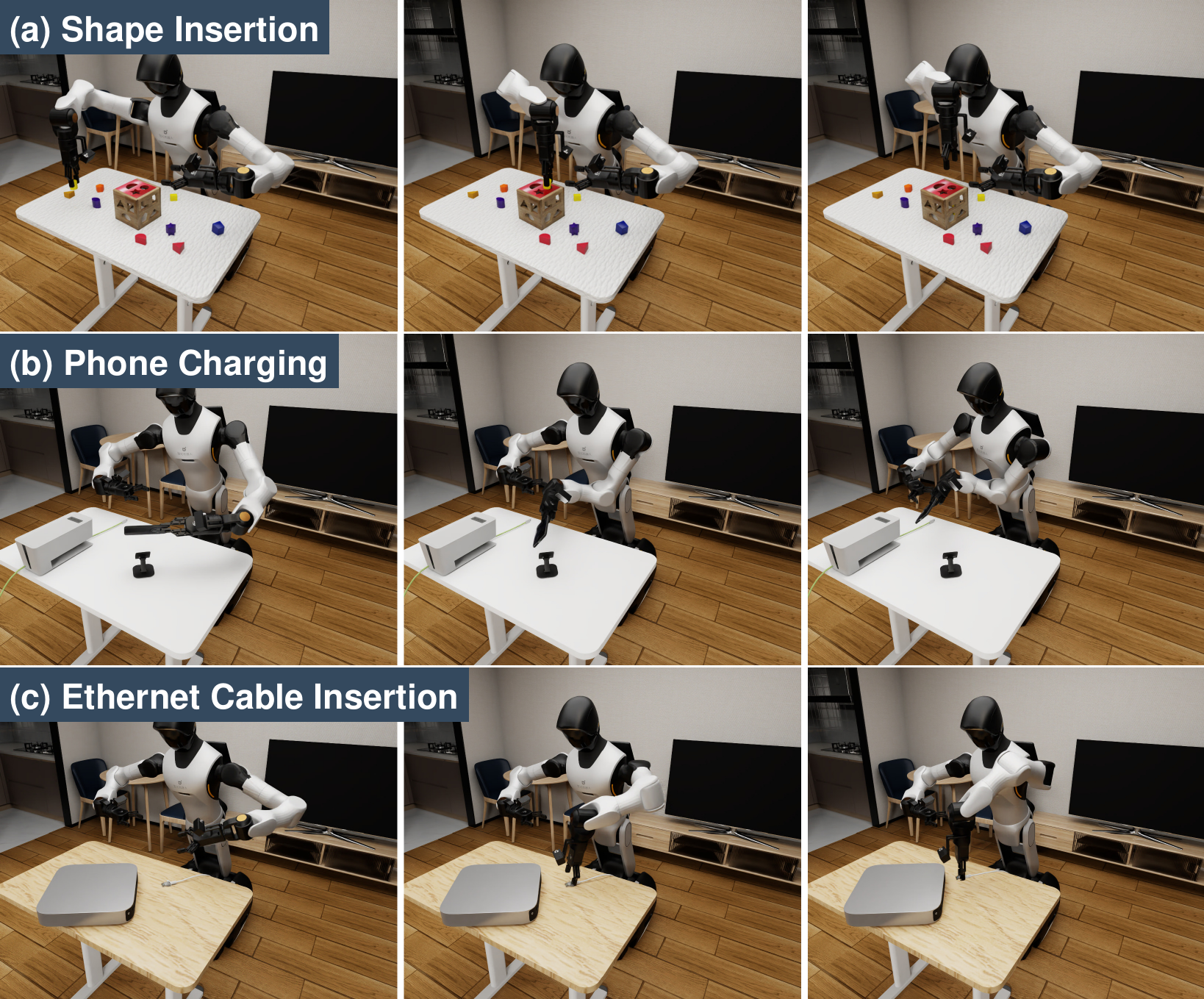}
\caption{Representative precision-manipulation sequences for
(a) block insertion into a matching hole, (b) phone charging, and
(c) Ethernet cable insertion. External-view frames progress
left to right, showing approach, grasping, and positioning
during the recorded attempts.}
\label{fig:case-precision}
\end{figure}

Shape insertion provides a successful example: after adjusting the
circular block's orientation, the model completes insertion. In phone
charging, the attempted alignment does not reach the charging port. In
Ethernet cable insertion, the connector is not aligned with the port
accurately enough to complete insertion. In coin
placement, the model misses the slot on its first attempt, leaving the
coin flat on the box lid; repeated subsequent attempts fail to pick it up
again. Figure~\ref{fig:case-precision} shows representative sequences.

\subsection{Manipulation Across Robot Embodiments}
To examine how the model's manipulation capabilities vary across robot
embodiments, we select four robot models: ARX, AgileX, G1, and G2. Each
embodiment uses its own kinematic model and tool geometry to convert
model-proposed end-effector targets into joint actions. Through the same
observation and action interface, the model completes tasks including
object pickup, desktop cleanup, and chess-piece storage. These successes
demonstrate its ability to manipulate objects across embodiments through
a common interface with embodiment-specific sensing and control adaptations.

\begin{figure}[!tp]
\centering
\includegraphics[width=\columnwidth]{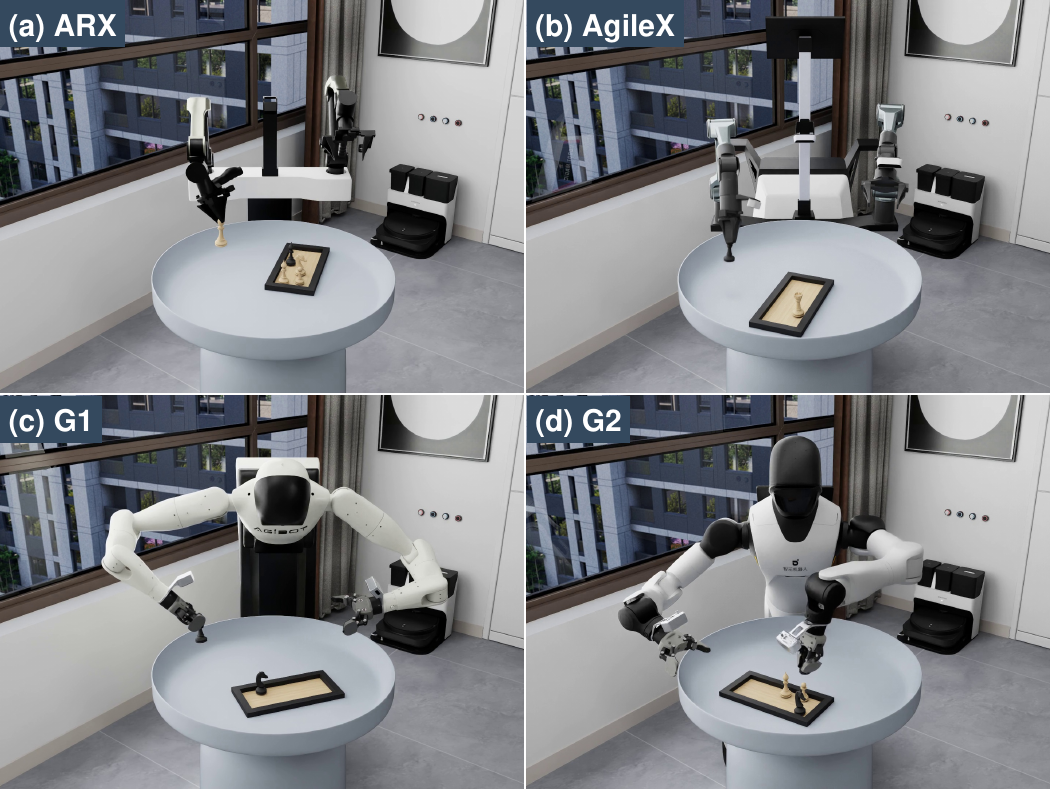}
\caption{Chess-piece manipulation on ARX, AgileX, G1, and G2 through the common end-effector interface.}
\label{fig:case-embodiments}
\end{figure}

Figure~\ref{fig:case-embodiments} illustrates chess-piece storage, which
the model completes on all four robots.

\subsection{Spatial Exploration and Mobile Interaction}
\begin{figure}[!tp]
\centering
\includegraphics[width=\columnwidth]{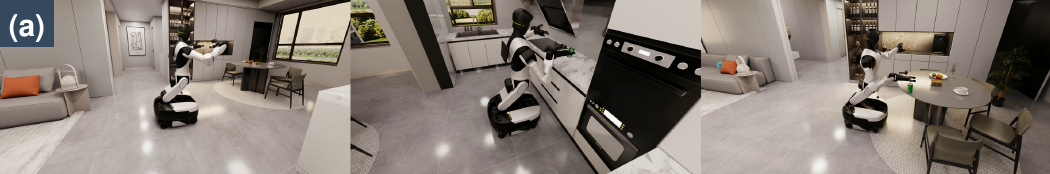}
\par\vspace{2pt}
{\fontsize{7}{9}\selectfont\textrm{\mbox{(a) Home: Get a Sprite from the kitchen and place it on the dining table.}}\par}
\vspace{5pt}
\includegraphics[width=\columnwidth]{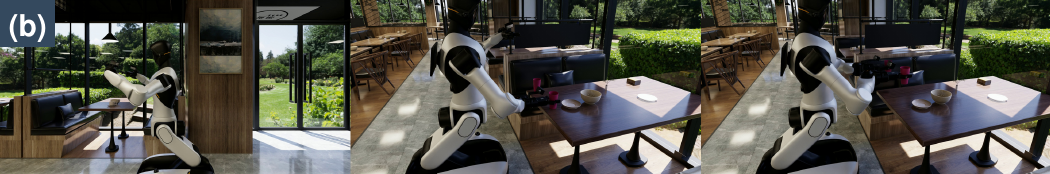}
\par\vspace{2pt}
{\fontsize{7}{9}\selectfont\textrm{\mbox{(b) Restaurant: Find and pick up pink cups from the dining table.}}\par}
\caption{Spatial exploration and mobile interaction in simulation.}
\label{fig:case-mobile-exploration}
\end{figure}

We construct diverse, open simulated environments to examine the model's
navigation, object search, and physical interaction capabilities
(Fig.~\ref{fig:case-mobile-exploration}). In the home task, the robot
identifies the kitchen, finds and retrieves a Sprite, and delivers it to
the dining table. In the restaurant task, it approaches the dining area,
locates the pink cups, and picks them up. These tasks demonstrate the
model's ability to combine spatial exploration, navigation, and object
interaction across multiple locations.

\subsection{Effects of Instruction Detail}
\label{sec:case-instruction-detail}
\begin{figure}[!tp]
\centering
\includegraphics[width=\columnwidth]{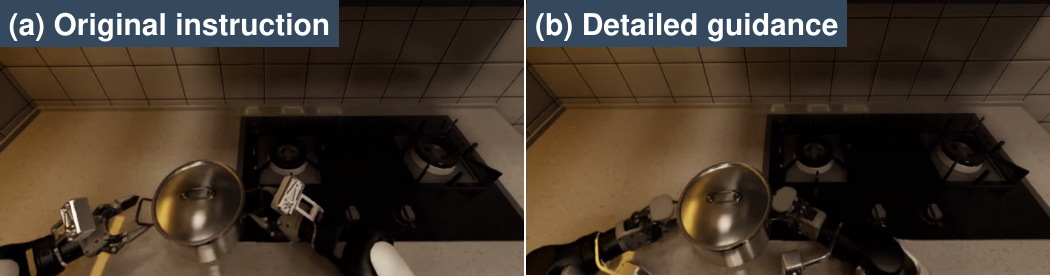}
\caption{Effect of instruction detail on pot holding.
(a, left) Poorly aligned grasp with the original task instruction.
(b, right) Better handle alignment and more stable handling with detailed
grasping and interaction guidance.}
\label{fig:case-pot-instruction-detail}
\end{figure}

The pot-holding case contrasts behavior under two levels of instruction
detail (Fig.~\ref{fig:case-pot-instruction-detail}). With the original
instruction, the model attempts a grasp with poorly aligned gripper
approaches. With explicit grasping and interaction guidance, it adopts a
better-aligned two-handed grasp at the handles and holds the pot more
stably.

Explicit interaction guidance improves the grasping strategy in this
case. Recognizing an object and understanding the task goal are not
always sufficient to select an appropriate physical interaction. This
highlights object-affordance understanding as a remaining challenge for
general manipulation.

\subsection{Real-Robot Manipulation on ARX Lift2}
\label{sec:real-robot-cases}
\begin{figure}[t]
\centering
\includegraphics[width=\columnwidth]{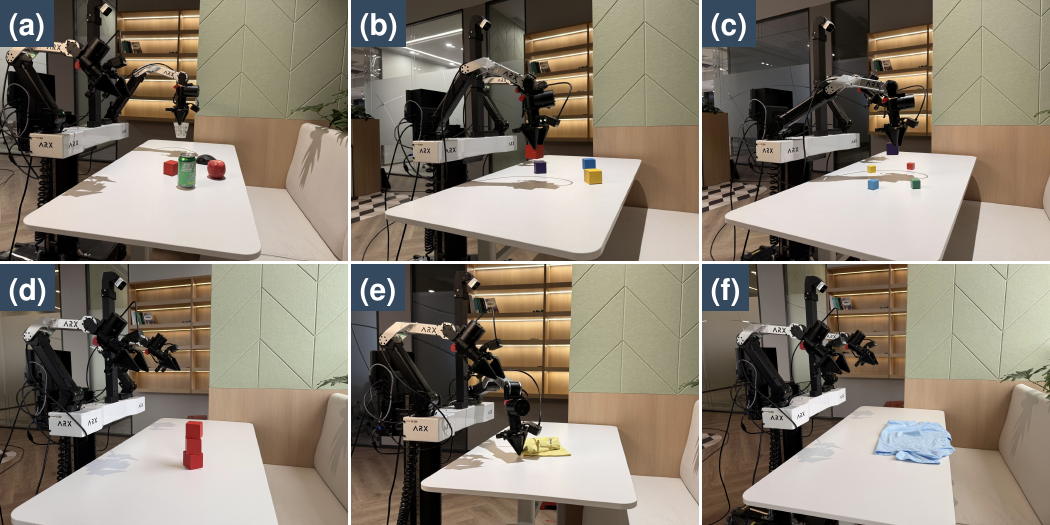}
\caption{Final recorded states of ARX Lift2 tasks in third-person views:
(a) cup pickup, (b) color-based block pickup, (c) size-based block pickup,
(d) block stacking, (e) towel folding, and (f) garment folding in half.}
\label{fig:real-robot-final-grid}
\end{figure}

We adapt the sensor and hardware-control interfaces to connect GPT-6 Astra
to an ARX Lift2 through Agent2Act, further exploring the model's ability to
control a real robot. Three onboard RealSense D405 cameras provide RGB
observations and depth information. Execution results are judged by human
observation.
Figure~\ref{fig:real-robot-final-grid} shows third-person views of the
final recorded task states. The trials highlight the following findings:

\begin{itemize}
\item \textbf{The model completes instruction-following, pick-and-place,
and stacking tasks on real hardware.}
Without depth input, the robot completes object selection, grasping,
placement, and block stacking, although execution is slow.

\item \textbf{Sensor depth supports contact-sensitive folding.}
Without depth input, towel folding and garment folding in half involve
repeated missed grasps and unsuccessful attempts. Both tasks succeed after
sensor depth is introduced, indicating that even noisy sensor depth can
help with precise contact positioning.

\item \textbf{Complex, multi-stage garment folding remains difficult.}
With depth input, the model grasps garment corners but fails to complete
the subsequent folding sequence. These attempts indicate that, with the
current task instructions, the model still struggles to interpret the
complex manipulation actions required for deformable objects.
\end{itemize}

\FloatBarrier
\section{Conclusion}
\label{sec:conclusion}
\label{sec:findings}
\label{sec:analysis}
\label{sec:future}
This report compares four frontier multimodal large language models on
RoboColiseum using the Agent2Act harness. Among these models, Opus leads
Overall, Spatial Reasoning, and General Manipulation, while Astra leads
Instruction Following and Robustness. No single model leads every dimension.
General manipulation remains a shared weakness: strong instruction
understanding and spatial reasoning do not yet ensure reliable complex
physical interaction. Comparisons on a limited test set highlight the
roles of observation inputs, feedback, history, and control interfaces in
task completion and execution efficiency.

Case studies with GPT-6 Astra further demonstrate whole-body coordination,
manipulation across robot embodiments, mobile navigation and interaction,
and task execution on real robots. Supplementary interaction guidance
improves how the model interacts with objects, but dexterous-hand control,
precision manipulation, and complex physical interaction remain substantial
challenges.

Despite the models' impressive capabilities across diverse embodied tasks,
inference cost and response latency remain the main barriers to practical
deployment.

We will continue to explore different multimodal models and agent systems,
characterizing their capability boundaries across robotic tasks,
embodiments, and environments. These findings will guide ongoing
improvements to RoboColiseum's evaluation capabilities, enabling model
capabilities and evaluation methods to co-evolve in an iterative feedback
loop.
\FloatBarrier
\nobalance

\clearpage
\nobalance
\onecolumn
\appendices
\section{Supplementary Results and Benchmark Task Renderings}
\label{app:ablations}
\label{app:benchmark-renderings}
\subsection{Task-Level Comparison Results}

\noindent\begin{minipage}{\textwidth}
\centering\begin{threeparttable}
\reporttabsetup
\renewcommand{\arraystretch}{1.2}
\captionof{table}{Task-level outcomes for the Instruction Following configuration comparisons (S: success; F: failure).}
\label{tab:ablation-tasks}
\small
\setlength{\tabcolsep}{4pt}
\begin{tabular*}{\textwidth}{@{\extracolsep{\fill}}p{0.30\textwidth}ccccccc@{}}
\toprule
\textbf{Task} & \textbf{\shortstack{Full\\system}} & \textbf{\shortstack{No\\feedback}} & \textbf{\shortstack{No\\depth}} & \textbf{Soft IK} & \textbf{\shortstack{No camera\\projection}} & \textbf{\shortstack{No gripper\\geometry}} & \textbf{\shortstack{Direct joint\\control}} \\
\midrule
pick\_billiards\_color & S & S & S & S & F & S & F \\
pick\_block\_color & S & S & F & S & S & S & F \\
pick\_block\_number & S & S & S & S & S & S & S \\
pick\_block\_shape & S & F & S & S & S & F & S \\
pick\_block\_size & S & S & S & S & S & S & S \\
pick\_common\_sense & S & S & S & S & S & S & S \\
pick\_follow\_logic\_or & S & S & S & S & S & S & S \\
pick\_object\_type & S & S & S & S & S & S & S \\
pick\_specific\_object & S & S & S & S & S & F & S \\
straighten\_object & F & F & F & S & F & F & S \\
\midrule
Success / episodes & 9/10 & 8/10 & 8/10 & 10/10 & 8/10 & 7/10 & 8/10 \\
\bottomrule
\end{tabular*}
\end{threeparttable}
\end{minipage}
\par\vspace{12pt}

\noindent\begin{minipage}{\textwidth}
\centering
\reporttabsetup
\renewcommand{\arraystretch}{1.2}
\captionof{table}{Full-system and no-history block-in-box outcomes and runtime (min; S: success; F: failure).}
\label{tab:history-layouts}
\small
\begin{tabular*}{\columnwidth}{@{\extracolsep{\fill}}cccrr@{}}
\toprule
 & \multicolumn{2}{c}{\textbf{Outcome}} & \multicolumn{2}{c}{\textbf{Runtime (min)}} \\
\cmidrule(lr){2-3}\cmidrule(l){4-5}
\textbf{Layout} & \textbf{Full} & \textbf{No history} & \textbf{Full} & \textbf{No history} \\
\midrule
0 & S & S & 80.76 & 65.00 \\
1 & S & F & 84.10 & 208.15 \\
2 & S & F & 46.90 & 199.41 \\
3 & S & S & 46.84 & 54.68 \\
\bottomrule
\end{tabular*}
\end{minipage}
\par\vspace{12pt}

\clearpage
\subsection{Benchmark Task Renderings}
The scene previews below illustrate task settings across the four
RoboColiseum tracks.

\begin{thebibliography}{99}

\bibitem{openai2026astra}
OpenAI, ``GPT-6 Astra: A new generation of intelligence,'' 2026.
\url{https://openai.com/index/gpt-6-astra/}

\bibitem{zhang2026unexpected}
W. Zhang et al., ``An Unexpected Robot Policy: Early Evaluations of GPT-6
Astra on RoboDojo and Beyond,'' arXiv:2609.24170, 2026.
\url{https://arxiv.org/abs/2609.24170}

\bibitem{su2026astra}
J. Su et al., ``GPT 6 Astra as an Embodied Policy,'' technical report, 2026.
\url{https://github.com/Zengyi-Qin/gpt-as-policy}

\bibitem{galbot2026embodied}
Galbot Team et al., ``Systematically Exploring the Capabilities of GPT-6
Astra as Embodied Policies,'' arXiv:2609.38537, 2026.
\url{https://arxiv.org/abs/2609.38537}

\bibitem{yao2026frontier}
Y. Yao et al., ``How Far Does a Frontier Agent Get as an Embodied Policy?,''
Shanghai Artificial Intelligence Laboratory, technical report, 2026.
\url{https://internrobotics.github.io/EBench-Astra-Tech-Report/}

\bibitem{zhang2026robocurve}
Z. J. Zhang, S. Machcha, S. Zou, T. K. Chan, and J. Chooi,
\emph{GPT-6 Astra vs MolmoAct2 on bimanual robotic manipulation},
RoboCurve, technical report, September 2026.
\url{https://openai.robocurve.org/stationerybench/}

\bibitem{liu2026roboicl}
F. Liu et al., ``RoboICL: Embodied In-Context Learning with GPT-6 Astra,''
arXiv:2609.34261, 2026.
\url{https://arxiv.org/abs/2609.34261}

\bibitem{openai2026sol}
OpenAI, ``Introducing GPT-6.1 Sol,'' 2026.
\url{https://openai.com/index/introducing-gpt-6-1-sol/}

\bibitem{anthropic2026opus}
Anthropic, ``Introducing Claude Opus 5.5,'' 2026.
\url{https://www.anthropic.com/claude-opus-5-5}

\bibitem{google2026argon}
Google, ``Gemini 4 Argon: our next era of frontier intelligence,'' 2026.
\url{https://blog.google/innovation-and-ai/models-and-research/gemini-models/gemini-4-argon/}

\end{thebibliography}
\end{document}